\documentclass[acmsmall]{acmart}

\usepackage{amsmath,amsfonts}
\usepackage{algorithmic}
\usepackage[ruled]{algorithm2e}
\usepackage{graphicx}
\usepackage{textcomp}
\usepackage{colortbl}
\usepackage{xcolor}
\usepackage[normalem]{ulem}
\usepackage{tabularx}
\usepackage{booktabs}
\usepackage{bm}
\usepackage{caption}
\usepackage{multirow}
\usepackage{mathtools}
\usepackage{stfloats}
\usepackage[utf8]{inputenc}
\usepackage[english]{babel}
\usepackage{enumitem}
\usepackage{url}
\usepackage{hyperref}

\usepackage{xcolor}
\usepackage{arydshln}
\AtBeginDocument{%
  \providecommand\BibTeX{{%
    \normalfont B\kern-0.5em{\scshape i\kern-0.25em b}\kern-0.8em\TeX}}}

\setcopyright{acmcopyright}
\copyrightyear{2022}
\acmYear{2022}
\acmDOI{10.1145/1122445.1122456}

\acmJournal{JACM}
\acmVolume{1}
\acmNumber{1}
\acmArticle{}
\acmMonth{1}

\begin{document}

\title{Adversary for Social Good: Leveraging Adversarial Attacks to Protect Personal Attribute Privacy}

\author{Xiaoting Li}
\affiliation{%
  \institution{Pennsylvania State University}
  \city{University Park}
  \state{PA}
  \postcode{16802}
  \country{USA}}
\email{xxl237@psu.edu}

\author{Lingwei Chen}
\affiliation{%
  \institution{Wright State University}
  \city{Dayton}
  \state{OH}
  \postcode{45435}
  \country{USA}}
\email{lingwei.chen@wright.edu}

\author{Dinghao Wu}
\affiliation{%
  \institution{Pennsylvania State University}
  \city{University Park}
  \state{PA}
  \postcode{16802}
  \country{USA}}
\email{dwu@psu.edu}






\begin{abstract}
  Social media has drastically reshaped the world that allows billions of people to engage in such interactive environments to conveniently create and share content with the public. Among them, text data (e.g., tweets, blogs) maintains the basic yet important social activities and generates a rich source of user-oriented information. While those explicit sensitive user data like credentials has been significantly protected by all means, personal private attribute (e.g., age, gender, location) disclosure due to inference attacks is somehow challenging to avoid, especially when powerful natural language processing (NLP) techniques have been effectively deployed to automate attribute inferences from implicit text data. This puts users' attribute privacy at risk. To address this challenge, in this paper, we leverage the inherent vulnerability of machine learning to adversarial attacks, and design a novel text-space \textbf{Adv}ersarial attack for \textbf{S}ocial \textbf{G}ood, called \textit{Adv4SG}. In other words, we cast the problem of protecting personal attribute privacy as an adversarial attack formulation problem over the social media text data to defend against NLP-based attribute inference attacks. More specifically, Adv4SG proceeds with a sequence of word perturbations under given constraints such that the probed attribute cannot be identified correctly. Different from the prior works, we advance Adv4SG by considering social media property, and introducing cost-effective mechanisms to expedite attribute obfuscation over text data under the black-box setting. Extensive experiments on real-world social media datasets have demonstrated that our method can effectively degrade the inference accuracy with less computational cost over different attribute settings, which substantially helps mitigate the impacts of inference attacks and thus achieve high performance in user attribute privacy protection.
\end{abstract}



\maketitle

\section{Introduction}
\begin{figure}[t]
	\centering
	\includegraphics[width=0.55\linewidth]{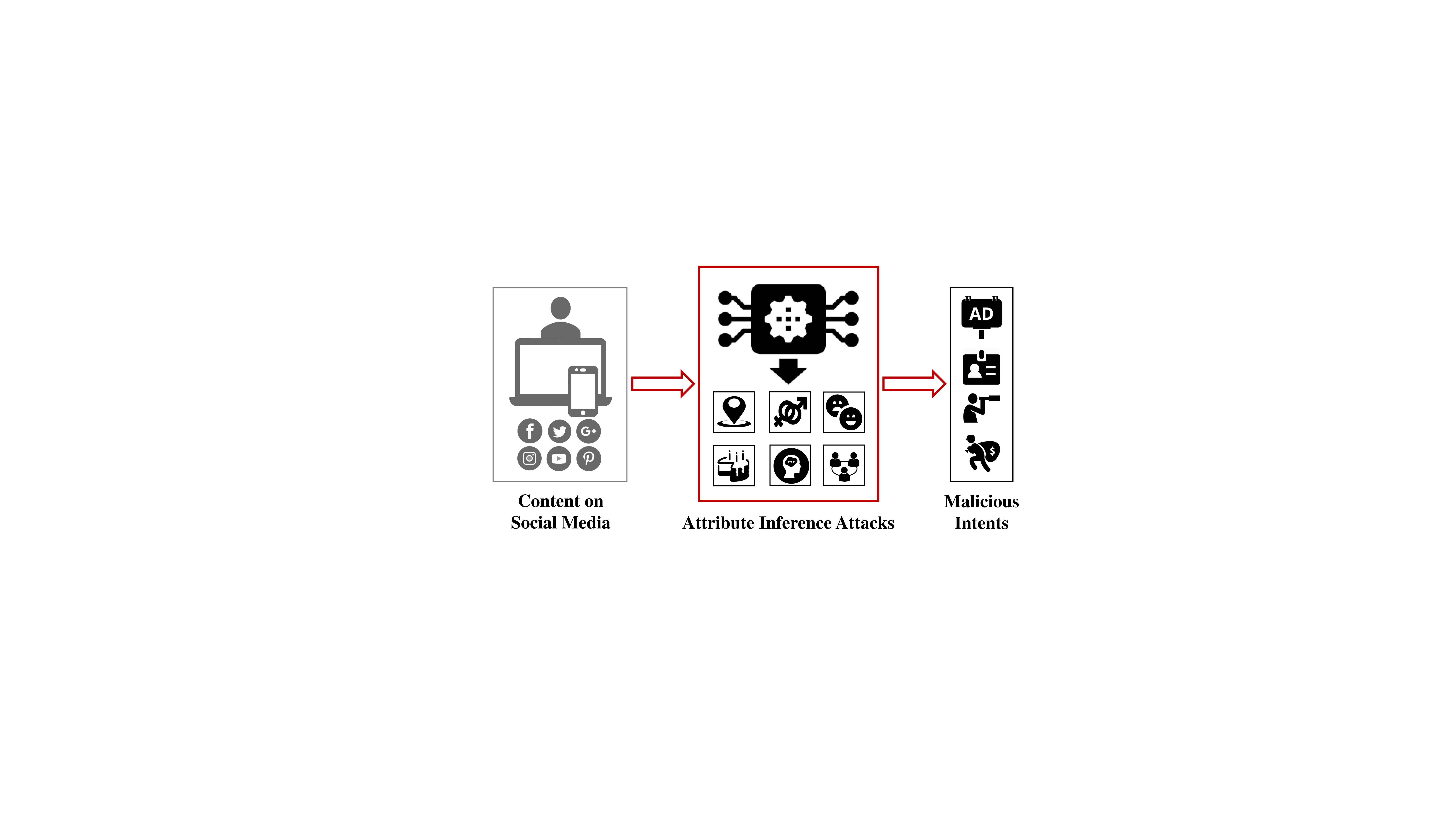}\\
	\caption{Attribute inference attacks over social media.} \label{fig:inference}
\end{figure}

In the Internet-age, social media undoubtedly has become an indispensable part of our daily lives through countless websites and apps, which allows us to discover and learn new information, create content and share ideas with friends, family and others. Such an interactive and convenient environment generates a mass of user-oriented data. Due to its accessibility and information richness, this data enables researchers to study and understand social communities and individual behaviors. For example, during the COVID-19 pandemic, a surge of solutions have been presented to leverage social media data for risk assessment \cite{ye2020alpha}. However, these apparent benefits also attract attackers to retrieve users' sensitive information and fulfill their malicious intents (e.g., unwanted advertising, user tracing) \cite{yu2018adversarial,beigi2018securing} as illustrated in Fig.~\ref{fig:inference}. Take Facebook data privacy scandal \cite{facebookscandal} as an example, the Cambridge Analytica harvested the personal data of millions of people from Facebook without their permission and used it for political advertising purposes. In fact, such privacy risk is not rare on social media, and could be quickly transmitted and propagated \cite{kumar2020adversary}.

Among the social media environment, text data maintains a huge amount of basic yet important user information. For example, users usually post their experiences, interests, comments or thoughts in the tweets or blogs for sharing. Such vibrant social engagements render text data a major target for attackers to parse the contents and reveal personal attributes (e.g., age, gender, location, sexual orientation, and political views) that people are unwilling to disclose. On the other hand, natural language processing (NLP) provides more and more powerful techniques for text understanding and mining, which enable 
a surge of effective attribute inferences from implicit text data that put social media privacy at risk \cite{morgan2017predicting,ikeda2013twitter,makazhanov2014predicting,ruder2016character,zhang2018tagvisor,gong2018attribute,jia2017attriinfer}. In this research work, we simply demonstrate an attribute privacy threat on social media as the scenario that an attacker trains a well-performed NLP model to infer users' private attributes from their text data such as tweets or blogs. With this in mind, some previous attempts have paid close attention to protect these attributes against inference attacks \cite{shokri2016privacy,wang2017locally,karadzhov2017case,jia2018attriguard,oh2017adversarial,shetty2018a4nt,jia2019memguard,kumar2020adversary}, which, however, still suffer from either large computational cost, or specific application scenarios limited to visual or high-dimensional data. Thus, our research goal here is to generalize the investigation into more challenging text data, and protect personal attribute privacy in this regard from a novel and practical adversarial learning perspective.

The effectiveness of machine learning models relies on the assumption that training data and test data follow the same underlying distribution, while this hypothesis is likely to be violated by an adversary who may manipulate the input data to compromise the output performance \cite{chen2017adversarial}. In other words, they are vulnerable to adversarial attacks that can easily fool the models into misclassification by adding small perturbations to the input data \cite{szegedy2013intriguing,goodfellow2014explaining}. Recent studies \cite{alzantot2018generating,li2018textbugger,gao2018black} have shown that NLP models are also faced with inherent learning-security challenge of lacking adversarial robustness. This naturally inspires us to take advantage of such a vulnerability and cast personal attribute privacy protection problem on social media as an adversarial attack formulation problem against attribute inference attacks. To achieve this goal, we face two challenges: (1) as inference attackers have a variety of choices in model construction, it is impossible for us to access the inference models in the real-world settings; (2) due to its discrete property and significant impact on social information interaction, modifications on text data have to comply with some essential constraints to guarantee the validity for the adversarial texts. 

To address the above challenges, in this paper, we first identify the practical black-box setting and the main types of constraints on text-space adversarial attacks, and then design an \textbf{Adv}ersarial attack for \textbf{S}ocial \textbf{G}ood, called \textit{Adv4SG}, to protect personal attribute privacy against NLP-based attribute inferences over social media text data. Given a source text (e.g., tweets, blogs), Adv4SG performs iterative word perturbations expedited by a reformed population-based optimization, in the sense that its target private attribute is misclassified by a self-trained NLP model. Through these adversarial perturbations, not only are the predefined text-space attack constraints enforced, but also the attribute obfuscation is very likely effected on the real attackers' inference models due to transferability in adversarial machine learning \cite{papernot2016transferability}. These advantages allow a refined paradigm to effectively mitigate the impacts of NLP-based inference attacks on attribute disclosure and enhance personal privacy protection in practical social media environment. In summary, our major contributions are listed as follows: 

\begin{itemize}
    \item A novel and practical paradigm of protecting personal attribute privacy on social media that leverages adversarial learning to mislead attribute inference attacks. 
    
    \item An adversarial attack is designed to obfuscate users' private attribute on more challenging text data of discrete property. Adv4SG is regulated by a reformed population-based optimization algorithm over perturbation subroutines that conform to text-space attack constraints, which can achieve better success rate in misclassifying attributes with less computational cost.
    
    \item The practical black-box setting is considered for Adv4SG's formulation, where the transferability of the proposed method is investigated to validate its applicability in real-world privacy protection scenarios.
    
    \item Extensive experimental evaluations on three real-world social media datasets (tweets and blogs) with different attributes to demonstrate the effectiveness of Adv4SG on attribute obfuscation and privacy protection.                    
\end{itemize} 

The rest of the paper is organized as follows. Section~\ref{sec:problemstatement} defines the problem of attack model for attribute inferences and adversarial attack for attribute protection. Section~\ref{sec:adversary} presents our detailed technical steps of text-space adversarial attack Adv4SG for attribute privacy protection on social media. Section~\ref{sec:experiments} evaluates the effectiveness of Adv4SG and the impact of different settings. Section~\ref{sec:limitations} discusses the applicability and limitations of our work. Section~\ref{sec:related} briefly introduces the related work. Section~\ref{sec:conclusion} concludes. 

\section{Problem Definition}\label{sec:problemstatement}
In this section, we first provide the problem definition of the attack model for attribute inferences, and then adversarial attack for attribute protection before technically detailing our proposed model Adv4SG in the following section. 

\subsection{Attack Model for Attribute Inferences} 

Social media enables users to post text data for social engagements. This data may bring privacy concerns to the forefront: the attackers that acquire such publicly exposed data may infer users' sensitive and private attributes (e.g., age, gender, and location) to deliberately fulfill the economic, social, or political intents, such as stealing user credentials, promoting unwanted advertisements, and stalking and threatening users \cite{yu2018adversarial,jia2018attriguard,beigi2018securing,shetty2018a4nt}. Considering that social media generally takes action to protect the explicit and identifiable information, in this paper, we assume that the attackers would train NLP models using the latent representations learned from the implicit information of text data to infer the attributes of interest. More specifically, we represent social media text data as ${\mathcal D} = \{d_{i}, y_{i}\}_{i = 1}^{n}$ consisting of $n$ sample texts, where each text $d \in {\mathcal D}$ is annotated with a ground-truth label $y \in {\mathcal Y}$ for a specific attribute. Taking location attribute (main four U.S. regions) as an example: ${\mathcal Y}$ can be accordingly specified as ${\mathcal Y} = \{0\text{:Northeast}, 1\text{:Midwest}, 2\text{:South}, 3\text{:West}\}$. We follow the general NLP routine to deal with discrete text data by mapping each text $d$ into a $k$-dimensional feature vector $\mathbf x = \phi(d)$ where $\phi$ is a feature representation function $\phi: {\mathcal D} \rightarrow \mathbf X \subseteq \mathbb{R}^{n \times k}$. In this respect, we can derive the predicted label of text $\mathbf x$ using the following formula:
\begin{equation}\label{eq:attributeinference}
    y^{*} = \underset{y \in {\mathcal Y}}{\arg\!\max}\ l_y(\mathbf{x})
\end{equation}
where $l_{y}(\mathbf x)$ is the confidence score of predicting sample text $\mathbf x$ as attribute label $y$ using an NLP model $l$ (e.g., convolutional neural network
(CNN), long short-term memory (LSTM), and Transformer). From Eq. (\ref{eq:attributeinference}), we can see that the final attribute label assigned to the input sample is the one with the highest confidence score.

\subsection{Adversarial Attack for Attribute Protection}

In the text space, we aim to design an adversarial attack to mislead attribute inference attacks and thus protect user attribute privacy. This is achieved in the way that the text-space adversarial attack perturbs the texts to obfuscate the target attribute and prevents inference attack models from correctly identifying their private attribute values.  
Specifically, given a text $\mathbf x$ and the associated attribute $y$ to protect, the formulated adversarial attack modifies the original text $\mathbf x$ to the adversarial text $\widehat{\mathbf x}$ by adding a small perturbation $\delta$, where $\widehat{\mathbf x}$ is predicted as any other label $\widehat{y} \in {\mathcal Y}$ ($\widehat{y} \ne y$). Therefore, we can formally define our objective function as follows:
\begin{equation}\label{eq:objective}
f(\mathbf x + \delta) = l_{y}(\mathbf x + \delta) - \max_{i\ne y}\{l_{i}(\mathbf x + \delta)\}
\end{equation}
Eq. (\ref{eq:objective}) distinctly indicates that $\mathbf x$ is misclassified as a member of $\widehat{y}$ if and only if $f(\mathbf x + \delta) < 0$ \cite{pierazzi2019intriguing}. 
The intuition to perform an adversarial attack in general feature space is to minimize $f(\mathbf x + \delta)$ by modifying $\mathbf x$ in the directions that follow the negative gradient of $f(\mathbf x + \delta)$ \cite{carlini2017towards,goodfellow2014explaining,papernot2017practical,moosavi2016deepfool}; that is, the adversarial attack can be implemented by solving the following optimization problem:
\begin{equation}\label{eq:optimization}
\begin{split}
    &\delta^{*} = \underset{\delta \in \mathbb{R}^{k}}{\arg\min}f(\mathbf x + \delta)\\
    &\text{\ \ s.t.\ \ } \|\delta\|_{p} < \epsilon \text{\ \ and\ \ } f(\mathbf x + \delta) < 0
\end{split}
\end{equation}
Due to its discrete property, these gradient-driven adversarial attack methods, however, cannot be directly applied to text space. The reasons behind this are that (1) gradients computed from the feature space are hard to define in text space, which may map the original text $\mathbf x$ to a set of non-admissible values; (2) $L_{p}$-norm distance metric typically works on continuous feature space, but is not capable of bounding the expected perturbation on texts represented as discrete tokens. Furthermore, a valid and realistic text-space adversarial attack for social good has to comply with some essential underlying constraints on the modification of the texts. These issues and challenges need to be addressed in the design of Adv4SG.

\section{Adversary for Attribute Privacy Protection}\label{sec:adversary}

In this section, we first identify the black-box setting and underlying constraints conformable to text-space attacks; guided by these constraints, we detail our adversary for social good idea of how we formulate an adversarial attack Adv4SG to protect attribute privacy against NLP-based inferences over social media text data. 

\subsection{Black-box Attack}\label{subsec:blackbox}

Considering the challenge that we are unable to access attacker's inference models, we put our work under the black-box setting, where the devised adversarial attack is not aware of any information about the inference model, including model choice, architecture, parameters, and training data. Compared to the assumptions made in \cite{alzantot2018generating,li2018textbugger,quiring2019misleading,papernot2017practical} that the attacks are able to retrieve the prediction scores by querying the target model with inputs, our black-box setting is more practical. In the real-world social media scenario, inference attackers have a variety of model choices, and it is impossible to specify one out of many. To this end, we self-learn a surrogate NLP model $l$ to perform attribute inference and craft adversarial texts. Similar to the attackers, we can train such an inference model using the public data and attribute values from the users. Due to transferability in adversarial machine learning \cite{papernot2016transferability}, the adversarial texts optimized to mislead the surrogate model are very likely to evade the real attackers' inference models.

\subsection{Text-space Attack Constraints}\label{subsec:constraints}

Different from the adversarial attacks in the general feature space, the generation of text-space adversarial attacks for social good is much more constrained. For example, small modifications on texts can be visually noticeable to human viewers, and lead to severe semantic loss on human understanding. In this respect, it is not feasible to obfuscate the attributes of a text through simply copying the words from another text with different attribute values for impersonation, or heavily manipulating the source text for evasion. Text-space adversarial attacks should thus comply with some essential constraints to guarantee their validity and applicability. In this section, we define these constraints as follows to guide our attack formulation and clarify its strengths.  


\vspace{0.1cm}\noindent\textbf{End-to-end learnability.} 
In order to generate a practical text-space adversarial text, the first and basic requirement to be achieved is the end-to-end learnability, which enforces iterative perturbations to be performed from text space to text space. In other words, the text-space adversarial attacks need to follow the transformation flow ${\mathcal D} \rightarrow {\mathcal D}$, where $d \mapsto \widehat{d}$ takes an original text $d$ and generates an adversarial version $\widehat{d}$. Since the feature representation function $\phi$ is generally not invertible, the challenge becomes to find a way to apply transformations $\delta$ on $d$ to generate $\widehat{d}$, so that $\phi(\widehat{d})$ is as close to $\widehat{{\mathbf x}}$ as possible \cite{kolosnjaji2018adversarial}. This suggests that the word perturbations on text $d$ should not be arbitrary, but guided by the misclassification of the target attribute. 

\vspace{0.1cm}\noindent\textbf{Visual similarity.} 
Modifications on texts are hard to be unnoticeable to human eyes. However, in order to increase the text validity and reduce the utility loss to facilitate its applicability in the social media environment, the generated adversarial texts should be perceptibly similar to the original ones as much as possible. This requirement can be satisfied by either perturbing the texts using the visually similar words, or restricting the number of words that are allowed to be modified.

\vspace{0.1cm}\noindent\textbf{Semantic preservability.} 
Preserving semantics is also one of the underlying requirements when generating high-quality adversarial texts in the context of social media. This indicates that the expressed semantic meanings from the original text $d$ and the adversarial text $\widehat{d}$ are required to remain consistent. In this regard, text- or word-level distance needs to be measured to guarantee the small difference caused by the perturbations preserves semantics for texts. On the text level, the edit distance (e.g., the number of perturbed words) between $d$ and $\widehat{d}$ constrained for visual similarity can help reach the semantic equivalence. On the word level, the Euclidean distance between the original and perturbed word embeddings can ensure the semantic similarity for each word transformation.

\vspace{0.1cm}\noindent\textbf{Text plausibility.} 
In addition to visual similarity, the text validity also requires the adversarial text is syntactically correct and readable to humans, which is considered as text plausibility. For our problem, text plausibility is important as the adversarial text would not only fool attribute inference attack models, but might also be posted on social media for displaying. For this reason, artifacts, which easily reveal that an adversarial text is invalid (e.g., garbled text, words with symbols), will not included. Note that, due to the fast-sharing and informal-writing property, user posts on social media may tolerate words with small misspellings or distortions to some extent, which are still valid and plausible to readers and social media.

\vspace{0.1cm}\noindent\textbf{Attack automaticity.} 
To be applied in practical use, the perturbations performed during the adversarial attack procedure need to be completely automated without human intervention. This requires that the possible and available changes made to the text $d$ exclude any transformations that are hand-crafted or need re-engineering on different datasets. In this way, the adversarial attack can be feasible to protect different attributes on different data scenarios without extra update efforts to the overall framework.

\begin{figure*}[t]
	\centering
	\includegraphics[width=\linewidth]{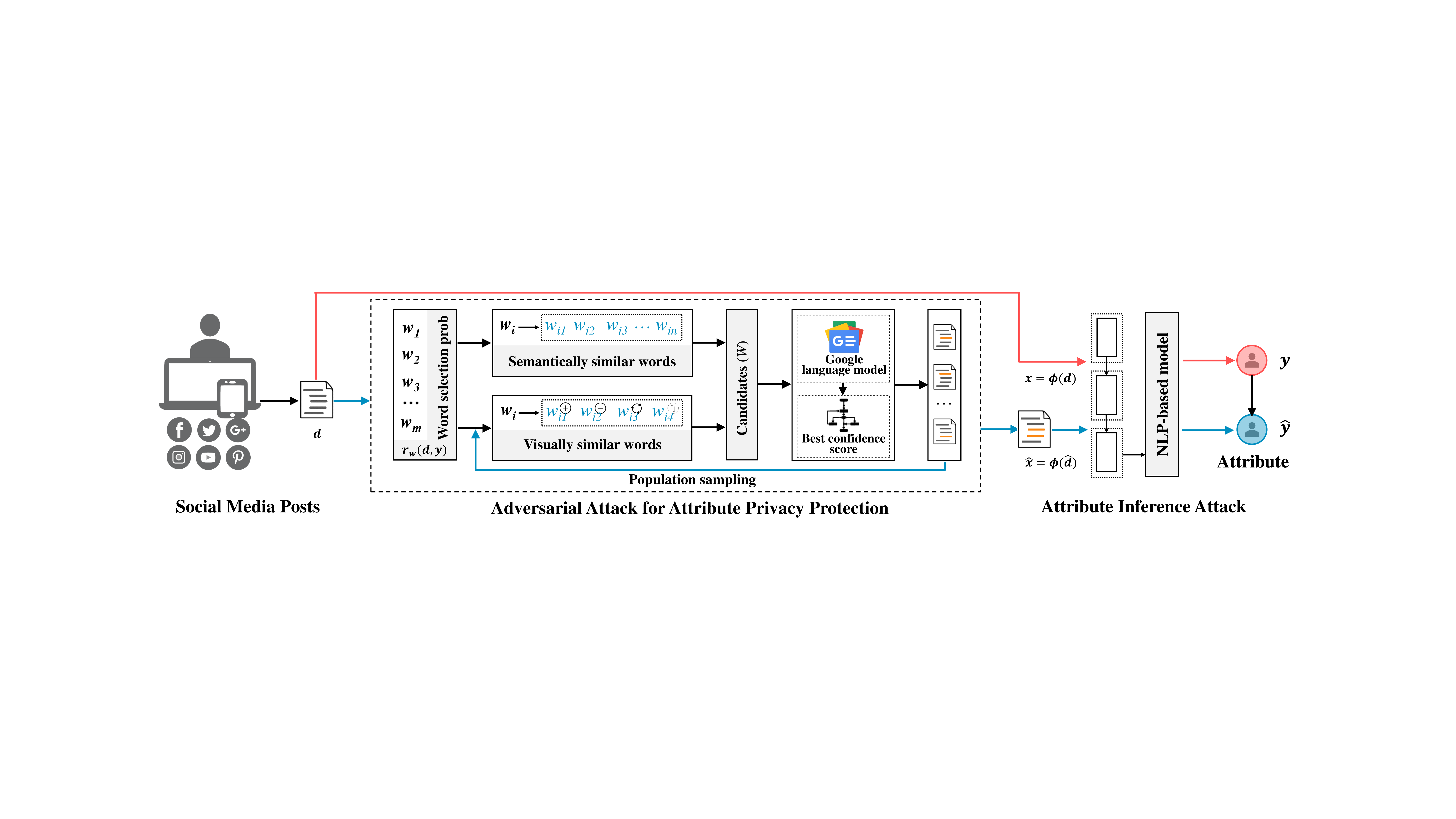}\\
	\caption{The overview of our proposed text-space adversarial attack Adv4SG for misleading inference attacks and protecting personal attribute privacy.} \label{fig:overview}
\end{figure*}

\subsection{Overview of Adv4SG}\label{subsec:overview}

The aforementioned real-world limitation and main types of constraints on text-space adversarial attacks raise significant challenges to the design of our attack method Adv4SG. To address these challenges, we propose Adv4SG to directly perturb the tokens in the text with guidance towards the misclassification of the target attribute through a self-trained NLP model, where the end-to-end learnability constraint and the black-box setting are naturally satisfied. Generally, tokens can be represented in the forms of words and characters, but in our attack formulation, we focus on perturbing the texts at word-level for two reasons: (1) the implicit information of the texts can be better encoded from the latent representations using word embedding than characters, which meets the assumption that the attackers would utilize the implicit information to train NLP-based models for attribute inferences; (2) the search space of possible changes over words is much smaller than characters, such that word-level perturbation is significantly more computationally tractable than character-level perturbation. Accordingly, we use edit distance metric in terms of the number of word changes to control the size of modifications so as to ensure the ability of fooling the threat model while remaining imperceptible. The overview of Adv4SG is illustrated in Fig.~\ref{fig:overview}. 

To this end, the feature-space adversarial attacks defined in Eq.~(\ref{eq:optimization}) can be updated to a text-space optimization problem as follows:
\begin{equation}\label{eq:newoptimization}
\begin{split}
    &\delta^{*} = \underset{\delta \in {\mathcal W}}{\arg\min}f(\phi(d + \delta))\\
    &\text{\ \ s.t.\ \ } \widehat{d} = d + \delta,\ \ s(\widehat{d}, d) < \epsilon \text{\ \ and\ \ } f(\phi(\widehat{d})) < 0
\end{split}
\end{equation}
where $s(\widehat{d}, d)$ denotes the number of different words between $\widehat{d}$ and $d$, ${\mathcal W}$ is the set of plausible and semantic-preserving word candidates for perturbation, and $+$ implies the high-level word change. From Eq.~(\ref{eq:newoptimization}), it is clear to see that Adv4SG proceeds with a sequence of word perturbations, where each perturbation takes the current text $d$, replaces a chosen word with the optimal candidate, and generates a new version $\widehat{d}$ such that $d$ and $\widehat{d}$ are semantically equivalent and visually similar, until the target attribute is misclassified or the maximum allowed perturbation $\epsilon$ is reached. It is worth remarking that, since all the operations and optimizations do not require manual intervention, and candidate constructions and word perturbations are defined and performed on the fly, we can accordingly ensure the automaticity for our attack.

\subsection{Perturbation and Optimization}\label{subsec:technicalsolutions}

For a text-space adversarial attack, it is significant to elaborate word perturbations and devise an effective optimization algorithm to guide the transformations towards the specified target \cite{quiring2019misleading}. Some existing works \cite{alzantot2018generating,gao2018black,li2018textbugger} have thus delivered promising results in adversarial text generation. Even so, there are still some downsides in these attack methods: (1) word perturbations are limited to either semantically similar candidate replacements or character transformations while ignoring each other; and (2) it is computationally expensive to find an optimal solution using greedy search or genetic algorithm with random population sampling. Differently, we advance Adv4SG by considering social media property, and introducing both semantically and visually similar word candidates for perturbations and a reformed population-based optimization to force attribute inference models to misbehave faster. We present the technical details of our proposed model Adv4SG in the following separate subsections.

\subsubsection{Scoring word importance}\label{subsubsec:scoring}

The original genetic attack proposed by Alzantot et al.\ \cite{alzantot2018generating} repeatedly performed perturbation on randomly selected word to formulate the population members at each generation, which may suffer from the vast search space of possible words and easily include those insignificant words. As such, we would like to first score the
importance of words in the text to guide the population sampling that touches the important words and thus expedite the adversarial text generation. 

Under our black-box setting, self-training NLP model allows us to compute the partial derivative of the confidence score regarding the predicted attribute label at each input word to obtain word saliency. Given the input text $d = (w_{1}, w_{2}, \cdots, w_{m})$, the scoring function that determines the saliency of $i$-th word in $d$ can be defined as:
\begin{equation}\label{eq:importance} 
    s_{w_{i}}(d, y) = \frac{\partial l_{y}(\phi(d))}{\partial w_{i}}
\end{equation}
where $l_{y}(\cdot)$ is the confidence score of predicting attribute label $y$. Based on our observation, a word's high saliency does not necessarily imply high importance if the perturbation performed on it fails to enforce high variation. Therefore, we further compute the perturbation variance on $i$-th word in the text $d$ as:
\begin{equation}\label{eq:importancetwo} 
        v_{w_{i}}(d, y) = \underset{\widehat{w} \in {\mathcal W}(w_{i})}{\arg\!\max}[l_{y}(w_{1}, w_{2}, \cdots, w_{n}) - l_{y}(w_{1}, \cdots, w_{i - 1}, \widehat{w}, w_{i + 1}, \cdots, w_{n})]
\end{equation}
where ${\mathcal W}(w_{i})$ is candidate set for $w_{i}$, which is constructed in Section~\ref{subsubsec:constructing}. Resting on the word saliency and perturbation variance, we approximate the importance of $i$-th word in the text $d$ as:
\begin{equation}\label{eq:importancethree} 
  r_{w_{i}}(d, y) = s_{w_{i}}(d, y) \cdot v_{w_{i}}(d, y)
\end{equation}

Clearly, the more important word has more impact on the model output, which is more likely to be modified to mislead inference model. Considering the facts that (1) some stop words (e.g., the, it, to, a, and an) or irrelevant words exist in a text that make little sense to tamper with and (2) the importance score of a word may be negative, we further use softmax function to normalize the word importance scores to serve as word selection probabilities for population sampling. In this way, the more important words in the sentences are given priority to be modified.

\subsubsection{Preparing word candidates}\label{subsubsec:constructing}

We focus on perturbing the texts at word-level; that is, we need to construct a set of word candidates for each selected word to perturb or replace. In order to satisfy the constraints that the generated adversarial text retains semantic equivalence and syntactic coherence with
the original one and visually imperceptible to human viewers on social media, we design two different types of word candidates for perturbation: semantically similar candidates and visually similar candidates. 

\vspace{0.1cm}\noindent\textbf{Semantically similar candidates.} We collect a set of words by searching the nearest neighbors of the ready-to-perturb word according to the Euclidean distance in word embedding space. To facilitate word search, a threshold $\eta$ is introduced to filter out candidates with distance greater than $\eta$ such that the semantic preservability requirement could be less violated. Compared to GloVe \cite{pennington2014glove}, Counter-fitting embedding \cite{mrkvsic2016counter} is a more context-aware word embedding space with fine-tuned semantic relations. Therefore, we use it to search for the nearest neighbors for the given word.

\vspace{0.1cm}\noindent\textbf{Visually similar candidates.} In addition to legitimate candidates from vocabulary, we also include the slightly perturbed words in the candidate pool. The reasons behind this choice are that (1) social media, as a fast-sharing and informal-writing environment, is highly misspelling-tolerant, where satiric or deliberate misspellings are not uncommon; (2) words with small character changes are imperceptibly to human eyes and have no significant impact on semantics \cite{rawlinson2007significance}, and (3) these words would very likely cause the selected word to be out of dictionary with ``unknown'' embedding such that the classification output may change \cite{gao2018black,li2018textbugger}. To ensure the text plausibility, we restrict that only small changes can be performed on the original word to create visually similar candidates, and those modified words will not be selected for a second perturbation. We design different word transformation methods as follows\footnote{Both the first and last positions in the original word will not be modified for better perturbation invisibility.}: 
\begin{itemize}
\item Add a space or a random character into the word except for the first and last positions.

\item Remove a random character from the word except for the first and last ones.

\item Swap any two adjacent characters except for the first and last ones.

\item Substitute a random character in the word with a randomly selected character except for the first and last ones.

\item Substitute a character or a substring to a visually (or aurally) similar number, such as $l \mapsto 1$, $o \mapsto 0$, $z \mapsto 2$, and straight $\mapsto$ str8. These are some deliberate formulations or slang on social media for user convenience or a rhetorical purpose.
\end{itemize}

\begin{algorithm}[t]
	\SetAlgoLined
	\SetKwFunction{KwPerturb}{PerturbationSubroutine}
	\SetKwProg{Fn}{Function}{:}{\textbf{end}}
    \Fn{\KwPerturb{$d$, $y$, $l$, $p$, $n$}}{
        $w$ = WordSelect($d$, 1, $p$)\;
        $\mathit{candsS}$ = SemanticConstructor($w$, $n$)\;
        $\mathit{candsV}$ = VisualConstructor($w$)\;
        \For{$c_{i} \in \mathit{candsS}+\mathit{candsV}$}
        {
            $d(i)$ $\leftarrow$ replace $w$ with $c_{i}$ in $d$\;
            $\mathit{score}(i)$ = $l_{y}(\phi(d(i)))$\;
            \If{$c_{i} \in \mathit{candsS}$}
            {
                $\mathit{pf}$, $\mathit{sf}$ $\leftarrow$ a word before/after $c_{i}$ in $d$\;
                $\mathit{gscore}(i)$ = GoogleLM($pf$, $c_{i}$, $sf$)\;
            }
        }
        $\mathit{tscore}$ $\leftarrow$ top $n/2$ in $\mathit{gscore}$\; 
        Remove $\mathit{score}(i)$ $\forall$ $c_{i} \in \mathit{candsS}$ and $c_{i} \not\in \mathit{tscore}$\;   
        $c = \arg\max_{c_{i}}\mathit{score}(i)$\;
        \KwRet $d(c)$\;
    }
	\caption{Perturbation subroutine.} \label{alg:algorithm1}
\end{algorithm}

\subsubsection{Selecting optimal candidate for replacement}\label{subsubsec:determing}

After collecting candidates for the word, we proceed by choosing optimal candidate to replace it. However, those constructed word candidates are not all feasible for selection, where some of the semantically similar candidates may not be used in the same context as others. For example, ``red'' and ``flushed'' are related neighbors, but obviously ``red'' can be widely used to depict anything red, while ``flushed'' more likely describes a face turning red. To address this issue, we pass all the semantically similar candidates through Google language model \cite{chelba2013one} to further filter out the ones that do not fit within the context and improve the semantic correctness. The rest are then integrated with visually similar ones to form the final candidates. Afterwards, we choose the optimal candidate among them that will maximize the confidence score of the target attribute $\widehat{y}$ ($\widehat{y} \ne y$) prediction when it replaces the ready-to-perturb word in $d$. Then we perturb the text with the optimal candidate and generate a new text as a population member.

\begin{algorithm}[t]
	\SetAlgoLined
	\KwIn{$d$: a text sample, $y$: label for a specific attribute, $l(\cdot)$: inference model, $\epsilon$: maximum perturbations, $n$: neighbor number.}
	\KwOut{$\widehat{d}$: an adversarial text.}
	\BlankLine
	Compute $r_{w}(d, y)$ using Eq. (\ref{eq:importancethree})\;
	$\mathit{selectprob}$ = Normalize$(r_{w}(d, y))$\;
	$\widehat{y}$ $\leftarrow$ label other than $y$\;
	${\mathcal P}^{0}$ = \{PerturbationSubroutine($d$, $\widehat{y}$, $l$, $\mathit{selectprob}$, $n$)\}$_{i=1}^{N}$\; 
	\For{$t = 1 \rightarrow I$}
	{
	    \For{$i = 1 \rightarrow N$}
	    {
	        $\mathit{score}(i)$ = $l_{\widehat{y}}(\phi({\mathcal P}^{t-1}_{i}))$\;
	    }
	    $p = \arg\!\max_{i}\mathit{score}(i)$,$\ $ $\widehat{d}$ = ${\mathcal P}^{t-1}_{p}$\;
	    \If{$s(d, \widehat{d}) \ge \epsilon$}
	    {
	       \KwRet None\;
	    }
	    \eIf{$\arg\!\max_{i}l_{i}(\phi(\widehat{d})) == \widehat{y}$}
	    {
	        \KwRet $\widehat{d}$\;
	    }
	    {
	        ${\mathcal P}^{t} = \{\widehat{d}\}$,$\ $ $\mathit{sampleprob}$ = Normalize$(\mathit{score})$\;
	        \For{$i = 2 \rightarrow N$}
	        {
	            $c$ = PopulationSampling(${\mathcal P}^{t-1}$, 2, $\mathit{sampleprob}$)\;
	            ${\mathcal P}^{t}$ = ${\mathcal P}^{t}$ $\cup$ PerturbationSubroutine($c$, $\widehat{y}$, $l$, $\mathit{selectprob}$, $n$)\;
	        }
	    }
	}
	\KwRet None\;
	\caption{Adv4SG for attribute privacy protection.} \label{alg:algorithm2}
\end{algorithm}

\subsubsection{Optimizing word perturbations}

The three steps detailed above can contribute to a \textit{perturbation subroutine} that accepts an input text (either perturbed or original), selects a word, perturbs it with optimal candidate, and generates a perturbed-version text towards the misclassification of the target attribute. The perturbation subroutine is illustrated in Algorithm~\ref{alg:algorithm1}. In this way, we are ready to generate a set of these perturbations for the given text. We aim to minimize the number of word perturbations, which makes the adversarial text more similar to the original one and less likely to be perceived. Therefore, instead of using greedy search \cite{gao2018black,li2018textbugger}, we follow the work by Alzantot et al.\ \cite{alzantot2018generating} and leverage a reformed population-based optimization algorithm to regulate the word perturbations during the formulation of Adv4SG.

The population-based optimization performs by sampling the population at each iteration, searching for those population members that achieve better performances, and taking them as ``parents'' to produce next generation \cite{alzantot2018generating}. This procedure can be summarized into three main operators. (1) Mutate($d$): select a word from the given input text $d$ using the normalized word importance score as the probability, and perform a perturbation subroutine on $d$. (2) Sample(${\mathcal P}$): sample a text $d_i$ from the population ${\mathcal P} = \{d_1, d_2, ..., d_N\}$ using the confidence score $l_{\widehat{y}}(d_i)$ as the probability. (3) Crossover($d_1$, $d_2$): construct a child text $c = (w_1, w_2, ..., w_m)$ where $w_i$ is randomly chosen from $\{w_i^{d_1}, w_i^{d_2}\}$. Based on these operators, population-based optimization first generates an initial population ${\mathcal P}^{0} = \{\text{Mutate}(d)_1, \text{Mutate}(d)_2, \text{Mutate}(d)_N\}$. At each following iteration $t$, the next generation of population will be generated in the following operation batch:
\begin{equation}
\begin{split}
    &\widehat{d}^{t} = \underset{d \in {\mathcal P}^{t-1}}{\arg\!\max}\ l_{\widehat{y}}(d),\\ &c_{i}^{t} = \text{Crossover}(\text{Sample}({\mathcal P}^{t-1}), \text{Sample}({\mathcal P}^{t-1})),\\
    &{\mathcal P}^{t} = \{\widehat{d}^{t}, \text{Mutate}(c_{1}^{t}), ..., \text{Mutate}(c_{N-1}^{t})\}
\end{split}
\end{equation}

The optimization will terminate when an adversarial text is found and returned, or the maximum allowed iteration number is reaches. Algorithm~\ref{alg:algorithm2} illustrates our proposed text-space adversarial attack Adv4SG. Different from the prior work, we improve the success rate of population samplings by choosing those ready-to-perturb words of high importance scores, while visually similar candidates introduced further expedite the adversarial text generation. Through Adv4SG, we can turn adversarial attacks into protection for personal attribute privacy on social media against the attribute inference attacks. 

\section{Experimental Results and Analysis}\label{sec:experiments}

In this section, we fully evaluate the effectiveness of our proposed adversarial attack Adv4SG for personal attribute privacy protection over social media text data.

\subsection{Experimental Setup}\label{subsec:experimentalsetup}

\vspace{0.1cm}\noindent\textbf{Datasets.}
We test our method on three real-world social media datasets: GeoText \cite{eisenstein2010latent}, user gender tweets\footnote{https://www.kaggle.com/crowdflower/twitter-user-gender-classification}, and blog authorship corpus \cite{schler2006effects}, which are good representatives for social media data as tweets and blogs are posted by different users, and easily accessed by attackers to uncover their private attributes. Specifically, GeoText is a tweet set from $9,500$ users with geographical coordinates in United States. We map each user into one of the main four U.S. regions defined by the Census Bureau\footnote{https://www2.census.gov/geo/pdfs/maps-data/maps/reference/us\_regdiv.pdf} and collect $9,281$ valid tweets with four locations (west, midwest, northeast and south). User gender tweets are collected from Kaggle. We filter out those with gender confidence score less than $0.5$, and obtain $13,926$ tweets with two genders (female and male). For blog data, it consists of $19,320$ documents, each of which contains the posts provided by a single user. We extract $25,176$ blogs with two attributes: (1) gender (female and male), and (2) age (teenagers (age between 13-18) and adults (age between 23-45)). Note that, age-groups 19-22 are missing in the original data. The data statistics are summarized in Table~\ref{tab:dataset}. 

\begin{table}[t]
    \centering
    \caption{\label{tab:dataset}Comparing statistics of the three datasets}
    \tabcolsep=4.5pt
    \begin{tabular}{ccccc}
    \toprule
          \textbf{Dataset} & \textbf{Attribute} & \textbf{\#Posts}  & \textbf{\#Classes} & \textbf{\#Vocabulary} \\
         \midrule
          Twitter\_g & Gender & 13,926 &2 &17k\\
          Twitter\_l & Location &9,281&4&16k \\
          Blog &Gender, Age & 25,176 &2&22k\\
    \bottomrule
    \end{tabular}
\end{table}

\vspace{0.1cm}\noindent\textbf{Text-space adversarial attack baselines.}
We compare Adv4SG with five other state-of-the-art text-space adversarial attack methods that are not only performed in an end-to-end manner at word level, but also representative to cover different formulations on word candidates and perturbation optimization. These attacks can be specified as follows:
\begin{itemize}
    \item Genetic attack \cite{alzantot2018generating}: this attack uses population-based optimization algorithm to generate adversarial examples with semantically similar candidates, where population sampling is performed in a random way at each generation.
    
    \item PSO attack \cite{zang2019word}: this attack uses sememe-based annotation method to craft word's substitution candidates and incorporates an adapted particle swarm optimization (PSO) strategy to search for adversarial examples.
    
    \item Greedy attack: this method greedily performs perturbation subroutine of our method on one word at each iteration. We aim to evaluate the performance of perturbation crafted by our subroutine and validate the effect of population-based optimization.
    
    \item WordBug \cite{gao2018black}: this attack scores word importance by removing it from text, and perturbs words in the descending order regarding word importance scores using character transformations. 
    
    \item TextBugger \cite{li2018textbugger}: this method also scores the word importance for greedy token selection, but proceeds by substituting the selected words with the optimal bug from candidates, including similar words in embedding space and word transformations.
\end{itemize}

\vspace{0.1cm}\noindent\textbf{Implementation details.}
We use euclidean distance as distance metric to construct semantic-similar candidates from embedding space, and the distance threshold is set to $\eta = 0.5$ to filter out those less similar ones. The size of candidate pool for each word is set as 8, where we choose the best one for replacement. We also limit the maximum allowed word perturbations to 25\% of the text length, and we further evaluate its impact on attack performance in Section~\ref{subsec:Adv4SGevaluation}. We randomly select 80\% of the samples for training, while the remaining 20\% is used for testing, and we report the mean inference accuracy and attack success rate of 3 runs on test samples for the evaluation results. For the system configuration, all the experiments are conducted on 2 $\times$ Intel(R) Xeon(R) Silver 4114 CPU with 512G RAM and 1 $\times$ TITAN XP 12GB.

\vspace{0.1cm}\noindent\textbf{Attack model for attribute inference attacks.}
An attribute inference attack aims to disclose private attributes of users by learning a model on the public data. Since we do not know the attacker's model, we self-train bidirectional LSTM (BiLSTM) \cite{graves2013generating}, multi-layer GRU (M-GRU) \cite{chung2014empirical}, ConvNets \cite{zhang2015character}, and CNN-LSTM (C-LSTM) \cite{gong2018adversarial} to perform the tasks. We mainly use BiLSTM to evaluate the effectiveness of Adv4SG, since it is one of the most popular and feasible neural networks to address NLP problems and can be easily built by the attackers to perform attribute inferences with relatively smaller cost, computing resource and training effort than transformer or BERT, which is more realistic in real-world inference attack scenarios. The comparisons among BiLSTM, M-GRU, ConvNets, and C-LSTM are leveraged for cross-model transferability evaluation in Section~\ref{subsec:transferability}. All models read in 250 words, where the dimension of each LSTM or GRU hidden unit is 128. We use GloVe \cite{pennington2014glove} to map each word into a 300-dimensional embedding space. Note that, an inference attacker would deploy more robust models to evade adversarial attacks. As adversarial training is considered as one of the most empirically robust methods against adversarial attacks \cite{athalye2018obfuscated,jia2019memguard}, we build up a robust model using adversarial training and further discuss the effectiveness of Adv4SG under this setting in Section~\ref{subsec:Adv_training}.

\subsection{Evaluation of Adv4SG}\label{subsec:Adv4SGevaluation}

In this section, we validate the effectiveness of Adv4SG against attribute inference attacks and the impacts of different parameters. To evaluate our method, we perturb the correctly classified text examples from the test data of four attribute settings. 

\begin{figure*}[t]
	\centering
	\begin{tabular}{c c c c}
	    \hspace{-0.3cm}
		\includegraphics[width=0.25\linewidth]{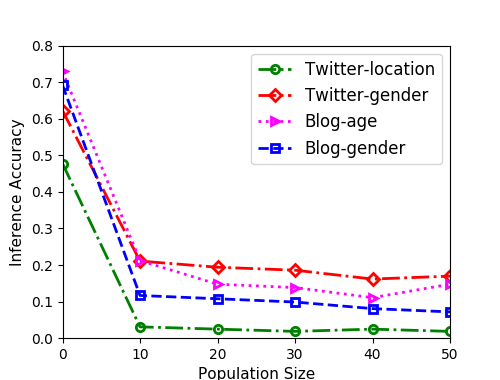}&
		\hspace{-0.3cm}
		\includegraphics[width=0.25\linewidth]{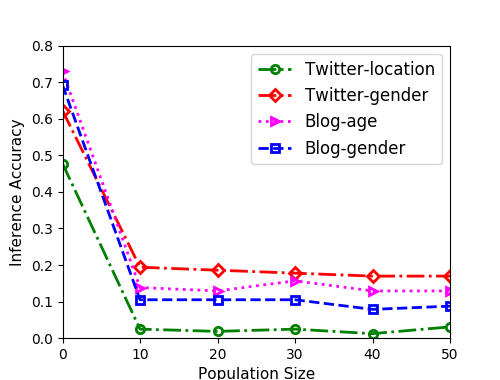}&
		\hspace{-0.3cm}
		\includegraphics[width=0.25\linewidth]{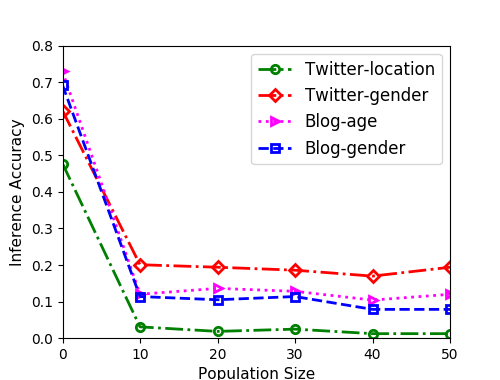}&
		\hspace{-0.3cm}
		\includegraphics[width=0.25\linewidth]{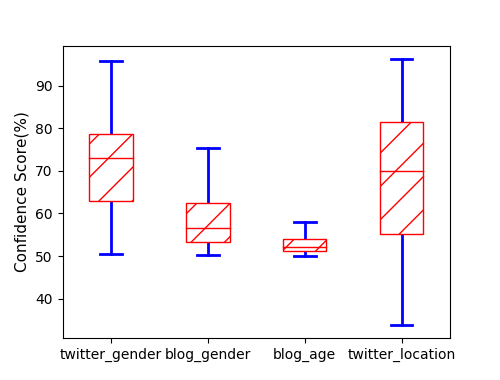}\\
		\textbf{\footnotesize (a) Iteration $I = 10$} & \textbf{\footnotesize (b) Iteration $I = 20$}&
		\textbf{\footnotesize (c) Iteration $I = 30$} & \textbf{\footnotesize (d) Score distribution}\\
	\end{tabular}
	\caption{Evaluation results: (a), (b) and (c) specify the inference accuracy of Adv4SG with different population sizes and iterations; (d) gives the confidence score distribution of the perturbed texts under four different inference settings.} \label{fig:evaluate}
\end{figure*}

\vspace{0.1cm}\noindent\textbf{Effectiveness.}
In our experiments, we evaluate Adv4SG under different population sizes and iterations as they play a crucial role to determine the degree of sample perturbation and computational cost. In particular, we test the results of our generated adversarial texts with population size $N \in \{10, 20, 30, 40, 50\}$ respectively against different inference attacks, while the maximum iteration $I$ is ranging in $\{10, 20, 30\}$ correspondingly. The experimental results are shown in Fig.~\ref{fig:evaluate}. As we can see from the results, the inference accuracy for Twitter-location, Twitter-gender, blog-age, and blog-gender on clean data is 47.76\%, 62.25\%, 72.92\%, and 69.20\%, which are relatively close to the state-of-the-art results on each dataset. Adv4SG drastically decreases all these accuracies and achieves the goal of obfuscating attributes and protecting social media text data privacy. Averagely, our method reduces the accuracy of Twitter-location and Twitter-gender inference attacks from 47.76\% to 2.19\% and from 62.25\% to 18.42\% respectively; for the larger and longer blog data, we degrade inference accuracy of gender and age from 69.20\% to 9.66\% and from 72.92\% to 13.65\% respectively. We present some of our generated adversarial texts in Fig.~\ref{fig:adv_cases}. It is clear that Adv4SG can subtly perturb important words towards the misclassification target in a plausible and semantic-preserving manner.

\begin{figure*}
    \centering
    \fbox{
        \begin{minipage}[tc]{0.95\textwidth}
            \begin{tabular}{@{}p{1\linewidth}@{}}
                \textbf{Task:} Twitter-location. \textbf{Original label:} South (confidence=76.88\%). \textbf{New label:} Northeast (confidence=61.66\%) \\
                \midrule
                They use the white \sout{\textcolor{red}{queso}} \textcolor{blue}{cheese} dip from farm fresh. I have seen cases of it in the kitchen. \\
            \end{tabular}

            \begin{tabular}{@{}p{1\linewidth}@{}}
                \midrule
                \textbf{Task:} Twitter-gender. \textbf{Original label:} Male (confidence=53.46\%). \textbf{New label:} Female (confidence=84.36\%) \\
                \midrule
                That \sout{\textcolor{red}{awesome}} \textcolor{blue}{amazing} moment when you \sout{\textcolor{red}{check}} \textcolor{blue}{chechk} your bank account and your parents send you more than you thought. \\
            \end{tabular}

            \begin{tabular}{@{}p{1\linewidth}@{}}
                \midrule
                \textbf{Task:} Blog-age. \textbf{Original label:} Adults (confidence=76.08\%). \textbf{New label:} Teens (confidence=60.31\%) \\
                \midrule
                Helloooooo! Well, in case you haven't guessed by the \sout{\textcolor{red}{lack}} \textcolor{blue}{l@ck} of my blogs, I have been on holiday \sout{\textcolor{red}{nowhere}} \textcolor{blue}{nowhare} nice just sitting at home. But I thought I \sout{\textcolor{red}{should}} \textcolor{blue}{shou1d} take a break from computers as well. I have lots of catching up to do, good news, bad news and lots of \sout{\textcolor{red}{events}} \textcolor{blue}{things} to tell you all about. So stay tuned for the updates!! \\
            \end{tabular}

            \begin{tabular}{@{}p{1\linewidth}@{}}
                \midrule
                \textbf{Task:} Blog-gender. \textbf{Original label:} Female (confidence=78.29\%). \textbf{New label:} Male (confidence=54.43\%) \\
                \midrule
                So it starts a \sout{\textcolor{red}{blog}} \textcolor{blue}{bl0g} on the internet ready for writing. I'm gonna \sout{\textcolor{red}{use}} \textcolor{blue}{utilize} this a lot over the \sout{\textcolor{red}{next}} \textcolor{blue}{future} two weeks to let you know what my theatre class is doing, the cute guys I'm meeting and all the rest enjoy. \\
            \end{tabular}
        \end{minipage}
    }
    \caption{Adversarial texts generated by Adv4SG under different inference tasks and their original texts.}
    \label{fig:adv_cases}
\end{figure*}

\vspace{0.1cm}\noindent\textbf{Impact of population size and iteration.}
Generally, when we enlarge the population size, the success rate of generating adversarial samples increases and the accuracy of the inference models thus decreases, while the required perturbation number tends to go up as well. However, due to the perturbation limit for each text, the actual attack performance might not always improve for larger population size. We can observe that the inference accuracy for all settings drops to the worst at $N = 40$ and then either slightly increases or stays flat when $N$ changes from 40 to 50. On the other hand, the larger iteration provides more improvement space for Adv4SG when the population size is small. For example, when $N = 10$, Adv4SG degrades the inference accuracy for blog-age setting from 21.23\% ($I = 10$) to 12.02\% ($I = 30$). Nevertheless, such inference accuracy difference among different iteration settings tends to be more statistically insignificant as the population size increases. As shown in Fig.~\ref{fig:evaluate}, Adv4SG achieves the comparable performance under all four inference settings at $N = 40$ with $I$ varying in $\{10, 20, 30\}$. The reason behind this is that the larger population size more likely enforces the optimal solutions at earlier iteration, while most of the failed population samples would stay in the loop at later iteration. Considering that the larger iteration may introduce more computational cost, while the larger population size can significantly enhance Adv4SG, we use $N = 40$ and $I = 10$ throughout the following evaluations to keep a good trace-off between the effectiveness and efficiency.  

\begin{figure}[t]
	\centering
	\includegraphics[width=0.65\linewidth]{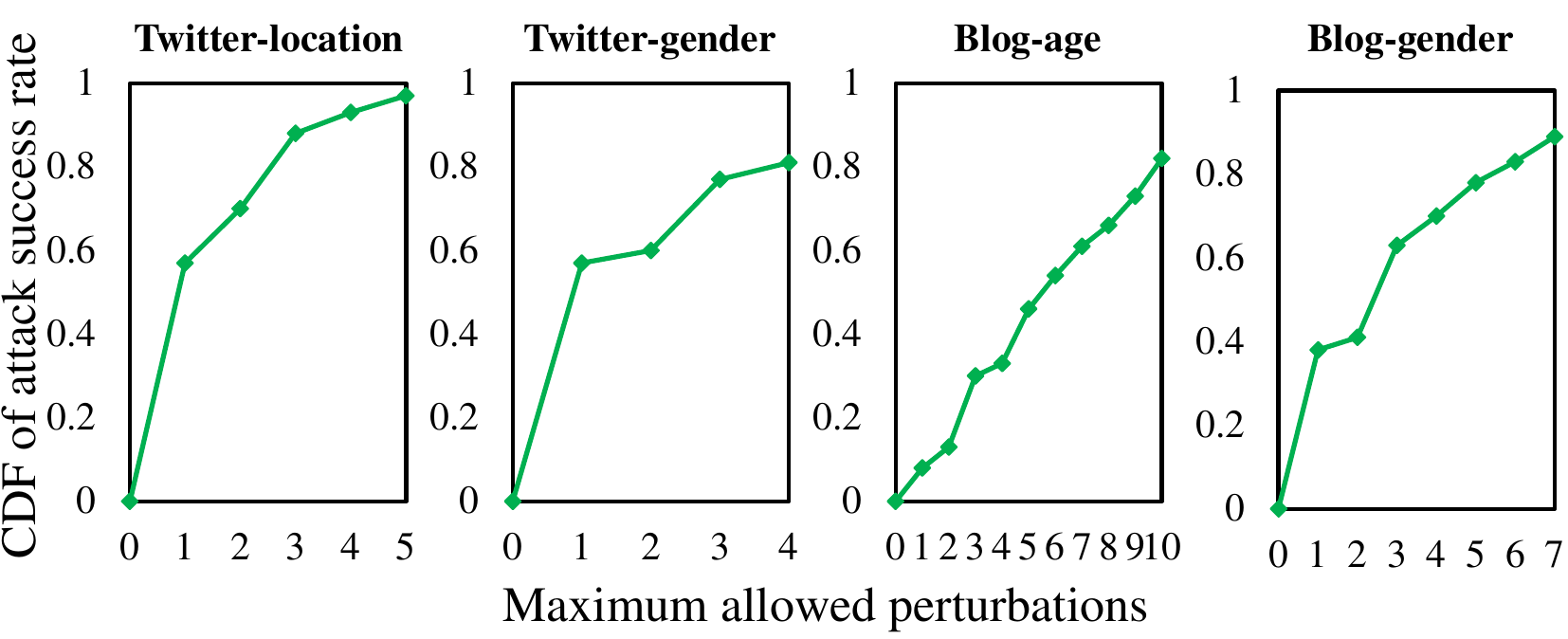}
	\caption{Evaluation on maximum allowed perturbation ($\epsilon$) via cumulative distribution of attack success rate.} \label{fig:parameter}
\end{figure}

\vspace{0.1cm}\noindent\textbf{Impact of maximum allowed perturbation ($\epsilon$).}
Different choices of $\epsilon$ could affect the performance of Adv4SG, since $\epsilon$ not only limits the number of word perturbations allowed to impact on the attack ability, but also significantly reflects the similarity between the generated adversarial texts and the original texts, and thus has direct impact on the semantic preservability and plausibility of the adversarial texts. We use the cumulative distribution function (CDF) of attack success rate regarding the number of $\epsilon$ to illustrate the evaluation results. From the results shown in Fig.~\ref{fig:parameter}, we can observe that as $\epsilon$ increases, the attack success rate increases as well because of the larger modification space, but the mean sentence semantics quality would decease. Actually, using Adv4SG, most of the generated adversarial texts manage to evade the inference models after perturbing very few words in the texts. More specifically, for Twitter-location inference, about 57\% of the testing texts evade the inference model by perturbing only one word, while this success rate increases to 88\% when $\epsilon \le 3$. For Twitter-gender inference, Adv4SG successfully crafts 57\% and 76\% of the adversarial texts from the original with at most one word and three word perturbations respectively. For blog-gender inference, the attack success rates are 38\% with $\epsilon \le 1$ and 63\% with $\epsilon \le 3$. For blog-age inference, these two rates are 9\% and 30\%, which apparently underperforms other settings because of the longer text length. When Adv4SG is allowed to perturb at most 5 words, the attack success rate immediately rises to over 50\%. All these results imply that (1) Adv4SG enables most of adversarial texts to be similar to the original texts; (2) the number of perturbations relatively relies on the length of the texts: the average lengths of the texts used for Twitter-location, Twitter-gender, blog-gender, and blog-age are 31, 15, 51 and 61, while the average perturbations are 1.8, 1.4, 2.9, and 5.2 for the corresponding inference tasks.

\vspace{0.1cm}\noindent\textbf{Other observations.}
In addition, we can also find some more interesting observations from the evaluation results in Fig.~\ref{fig:evaluate} and Fig.~\ref{fig:parameter}: (1) Adv4SG tends to perform worse on binary attributes (e.g., age and gender) than multi-class attributes (e.g., location). It is not difficult to understand that adversarial attacks on binary attributes can be considered as targeted attacks that might take more effort to perturb the texts and enforce misclassification to a specified target class (inverse to the original), while adversarial attacks on multi-class attributes fall into non-targeted attacks that have to simply cause the source texts to be misclassified, which is obviously easier. (2) The learning ability of the inference model may also have a potential impact on the Adv4SG's attack effectiveness against it, as small perturbations on the texts more likely lead to evasion for inference models that underperform than others. For example, the inference accuracy for Twitter-location is 47.76\%, while Adv4SG successfully reduces it to 2.19\% with 7.65\% mean perturbation rate. The similar results can be found between blog-age and blog-gender. (3) The age attribute seems more difficult to be obfuscated than others due to relatively higher model inference ability and longer text length, where Adv4SG performs more word perturbations for adversarial text generation.

Furthermore, we show the confidence distributions of those generated adversarial texts that can successfully fool the inference attackers under different deployment settings in Fig.~\ref{fig:evaluate}(d). It indicates the consistent findings with what we observe from other results. For instance, the average confidence values of the perturbed texts for the age attribute are distributed slightly above the borderline (i.e., 50\%), which reveals the difficulty in obfuscating age attribute for blog dataset. Differently, the overall scores of other three tasks have been explicitly moved to the misclassification direction, which lead to better attack effectiveness. In addition, the performance of Adv4SG for long texts (i.e., blogs) seems to be more stable than short twitter texts. We guess it correspondingly relates to the different inference capability of the attackers on these datasets. 

\begin{table*}[t]
	\centering
	\small
	\caption{Comparisons of different text-space adversarial methods}\label{table:comparison}
	\tabcolsep=3.5pt
	\begin{tabular}{llcccccc}
		\toprule
		\textbf{Inference task} & \textbf{Metric} & \textbf{Adv4SG} & \textbf{Genetic} &\textbf{PSO} & \textbf{Greedy} & \textbf{WordBug} & \textbf{TextBugger} \\
		\midrule
		\multirow{3}{*}{\textbf{Twitter-location} }&Success Rate & \textbf{97.40\%}&85.71\%&94.70\% &76.62\%&55.84\% &82.91\% \\
		                        &Median Ptb Rate & \textbf{5.26\%}&6.25\%&5.79\% &8.33\%&10.53\%& 7.85\%\\
		                        &Mean Ptb Rate & 7.65\%&9.00\%&\textbf{7.33\%} &10.73\%&18.75\% &11.58\% \\
		\midrule
		\multirow{3}{*}{\textbf{Twitter-gender} }&Success Rate & \textbf{74.03\%}&55.84\%&67.80\% &45.45\%&32.47\%& 62.34\%\\
		                        &Median Ptb Rate & \textbf{9.09\%}&14.29\%& 10.24\%&14.64\%&27.27\%& 16.67\% \\
		                        &Mean Ptb Rate & \textbf{12.18\%}&16.28\%& 12.72\%&16.73\%&29.56\% &21.37\% \\
		\midrule
		\multirow{3}{*}{\textbf{Blog-age}}&Success Rate &\textbf{82.28\%}&72.15\%& 74.05\%&72.15\%&17.72\%& 59.49\% \\
		                        &Median Ptb Rate &11.92\%&\textbf{11.11\%}& 12.90\%&12.19\%& 31.21\%&19.64\% \\
		                        &Mean Ptb Rate & \textbf{13.53\%}&13.96\%& 14.44\%&14.06\%& 27.94\%&23.89\% \\
		 \midrule
		\multirow{3}{*}{\textbf{Blog-gender}} &Success Rate &88.61\%&84.81\%& \textbf{88.87}\%&70.89\%&54.43\%&77.22\% \\
		                        &Median Ptb Rate &5.08\%&\textbf{4.21\%}& 4.85\%& 7.45\%& 17.86\%&12.31\% \\
		                        &Mean Ptb Rate & 8.38\% &8.61\%& \textbf{8.14}\%& 10.33\%& 19.07\%&16.03\% \\
		\bottomrule
	\end{tabular}
\end{table*}

\subsection{Comparisons with Other Attack Baselines}\label{sec:compare}

\vspace{0.1cm}\noindent\textbf{Attack performance.}
We compare Adv4SG with the other baselines including Genetic attack \cite{alzantot2018generating}, PSO attack \cite{zang2019word}, Greedy attack, WordBug \cite{gao2018black}, and TextBugger \cite{li2018textbugger}. These attacks are composed of word candidate preparation and perturbation optimization, but we follow the formulations presented in the related works and compare with them as a whole. Note that, in the original design of PSO attack, all words with the same sememe annotations serve as the perturbation candidates for the given word, the quantity of which could be very large on average. Considering that the candidate pool size of each word is set as 8 for all the other attacks, we narrow down the size of word candidates as the substitutes to 8 as well during PSO search for fair comparisons. From another perspective, we can also enlarge the candidate pool size to further elevate Adv4SG's attack performance, since each given word could have derived many more accessible neighbors under the distance threshold $\eta = 0.5$, let alone relaxing this restriction. As such, this experimental setting does make sense. To perform the evaluations, we randomly sample 50\% of correctly classified examples from the testing tweets and blogs to measure the performance of attacks. 

The comparative results are reported in Table~\ref{table:comparison}, where Genetic attack outperforms Greedy attack, WordBug, and TextBugger, while TextBugger performs slightly better on tweet attribute obfuscation; PSO attack achieves higher attack success rate and perturbs less words than Genetic attack; Adv4SG outperforms PSO attack in most settings with marginally lower attack success rate and higher perturbation rate on blog-gender inference. From the results, we can observe that (1) PSO with trade-off between local exploration and global exploitation is a more effective optimization method than genetic algorithm, while reducing the sememe-based word candidate size would greatly degrade the attack performance and yield less advantages out of particle swarm search; (2) projecting an important word into ``unknown'' may enforce inference models to misbehave faster, while ignoring semantically similar candidates would also miss good evasion chances, and (3) leveraging word importance to facilitate population-based optimization advances and expedites adversarial example generation. When we look into the generated adversarial texts, we find that Greedy attack fails in some of those adversarial texts with more modifications required over long blogs, and hence obtains a smaller perturbation number on average in results. By contrast, Adv4SG either converts those failed texts to adversarial examples, decreases the number of required perturbations, or raises the confidence scores of the perturbed texts, which refines the text-space adversarial attack with respect to effectiveness and efficiency. Thus, Adv4SG can be a feasible paradigm in a real social media environment on attribute obfuscation and privacy effectiveness.

\begin{figure}[t]
	\centering
	\includegraphics[width=0.48\linewidth]{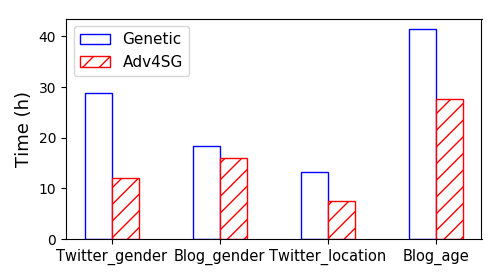}\\
	\caption{Computational cost between Adv4SG and Genetic.} \label{fig:time}
\end{figure}

\vspace{0.1cm}\noindent\textbf{Computational cost.}
We evaluate the computational cost in terms of running time. On one hand, the greedy-based attack methods (i.e., Greedy, WordBug, and TextBugger) follows the heuristic of making the locally optimal choice and perturb one word at each stage; hence their search space is much smaller than genetic algorithm and their running time is undoubtedly much smaller as well, but the greedy methods fail to generate successful adversarial examples on a larger number of texts which drastically underperform genetic attacks. Also, due to the guidance of individual best position, global best position and the corresponding update probabilities, PSO attack manages to explore more positions and find the adversarial texts more quickly than genetic optimization, but requires a significantly large amount of preprocessing time to perform sememe annotations, sense identification, and word candidate generations before perturbation optimization, which is very different from genetic attack formulation. On the other hand, our proposed adversarial attack Adv4SG extends the original genetic design to advance the evolution process towards better solutions and expedite the word perturbations. Thus, here we would like to merely evaluate the runtime advancement of our method against genetic method. 


To be comparable, we use single TITAN Xp for each experiment. We measure the average runtime for different inference settings on Genetic and Adv4SG, respectively. The results are presented in Fig.~\ref{fig:time}. We can see from the results that Adv4SG can drastically reduce the running time compared to Genetic. For inference tasks such as Twitter-gender and Blog-age, Genetic method costs nearly twice the time of our method. On average, Adv4SG can save 36.85\% computational time against the Genetic, which further justifies the advantage of word importance and visually similar candidates introduced in our proposed attack model Adv4SG.


\subsection{Decomposition Analysis}\label{subsec:decomposition}
In this section, we conduct a decomposition analysis to investigate how different components impact on the performance of our method Adv4SG with respect to attack success rate. In our model design, Adv4SG proceeds with word candidate preparation including semantically similar and visually similar candidates, and perturbation optimization that follows the genetic algorithm \cite{alzantot2018generating} to advance the population-based optimization to be more effective and efficient for adversarial text generation. Therefore, to verify their contributions, we separate Adv4SG into these components and analyze their contributions to the attack performance from formulating four alternative models: (1) \textit{Semantic+Genetic}: only semantically similar candidates are crafted based on the vector distance in counter-fitting embedding space for word substitution, and the genetic algorithm is deployed for population-based optimization to find adversarial examples; (2) \textit{Visual+Genetic}: this model prepares visually similar candidates for each word by using the designed transformation methods, and incorporates genetic optimization to regulate the adversarial text generation. (3) \textit{Semantic+Visual+Genetic}: we collect both semantically similar candidates and visually similar candidates for word substitutions, where word perturbations are again optimized using genetic algorithm. (4) \textit{Adv4SG}: the complete design of our attack model. 



\begin{table}[t]
    \centering
    \caption{Evaluation on different attack combinations with respect to attack success rate}\label{tab:ablation}
    \vspace{-0.2cm}
    \tabcolsep=5.5pt
    \begin{tabular}{lcccc}
        \toprule
         Method & Twitter-location& Twitter-gender & Blog-gender & Blog-age  \\
        \midrule
        Semantic+Genetic &85.71\%&55.84\%&72.15\%&84.81\% \\
        Visual+Genetic &58.75\%&44.16\%&50.10\%&56.96\% \\
        Semantic+Visual+Genetic &92.40\%&71.73\%&78.48\%&86.87\%\\
        Adv4SG (Ours) &\textbf{97.40\%}&\textbf{74.03\%}&\textbf{82.28\%}&\textbf{88.61\%}\\
        \bottomrule
\end{tabular} 
\end{table}

The experimental results for decomposition analysis are reported in Table~\ref{tab:ablation}. From the results, we can observe that when substituting words using individual candidate set, semantically similar candidates achieve better results than visually similar candidates as the latter substitutions can merely provide ``unknown'' embedding space to impact on the text semantics, while semantically similar candidates may enable better evasion chances against the inference model with more diverse and dynamic perturbation possibilities to explore. After putting these two candidate sets together, visually similar candidates surprisingly play a crucial role in adversarial text generation, where the attack success rate increases by 6.69\%, 15.89\%, 6.33\%, and 2.06\% for four inference tasks, respectively. This further confirms that enforcing the chosen words into ``unknown'' may mislead the inference model in a faster way, which provides a ``shortcut'' to gather around the optimal positions based on the early effort made by word perturbations using semantically similar candidates. The attack performance discrepancy between Semantic+Visual+Genetic and Adv4SG demonstrates that population sampling guided by word importance is able to further advance the
state-of-the-art performance to a higher level, which implies that this operation yields an additional advantage for population-based optimization that random population sampling may have missed. These observations from decomposition analysis highlight the effectiveness of Adv4SG. 


\subsection{Transferability}\label{subsec:transferability}
Under the black-box attack setting, as Adv4SG is implemented through self-trained NLP model, it is necessary to evaluate its transferability to validate if those adversarial texts generated for one model are likely to be misclassified by others. In this evaluation, we deploy Adv4SG to generate adversarial texts on four inference settings for four different NLP models: BiLSTM \cite{graves2013generating}, multi-layer GRU (M-GRU) \cite{chung2014empirical}, ConvNets \cite{zhang2015character}, and CNN-LSTM (C-LSTM) \cite{gong2018adversarial}. Then, we evaluate the attack success rate of the generated adversarial texts against other models. To ensure our results are comparable, we build up these models with the same parameter settings (different dropout rates) and training data. Accordingly, we build a cross-model transferability table, where each table unit $(i, j)$ holds the percentage of adversarial texts crafted to mislead model $i$ (row index) that are misclassified by model $j$ (column index). 

From Table~\ref{tab:transfer}, we can see that the cross-model transferability for Adv4SG is a strong but heterogeneous phenomenon: (1) between same model pairs, the percentage numbers are higher than 80\%, most of which are close or beyond 90\%; (2) between pairs of different models, some enjoy good transferability (e.g., 76.67\% for M-GRU and BiLSTM on blog-gender setting), while some only have moderate one (e.g., 31.03\% for ConvNets and M-GRU on blog-age setting). The results also imply that the complexity of the surrogate model and the intrinsic adversarial vulnerability of the target model contributes to attack transferability (e.g., all models against ConvNets achieve relatively higher transferability than others). Adversarial texts generated from more complicated surrogate model tends to have better attack success rates on other target models. We believe it is because models with complex structures enjoy high capability of regularization on malicious perturbations wherefore adversaries need to enlarge the input mutations to fool the model. In real-world scenarios, since the target models are uncontrollable and inaccessible, social media may need to elaborate the surrogate model for better transferability when applying Adv4SG for attribute privacy protections.

\begin{table*}[t]
    \centering
    \tiny
    \caption{Transferability on four inference settings: each unit $(i, j)$ specifies the percentage (\%) of adversarial texts produced for model $i$ that are misclassified by model $j$ ($i$ is row index, while $j$ is column index)}\label{tab:transfer}
    \tabcolsep=1.1pt
    \begin{tabular}{lcccccccccccccccc}
    \toprule
    \multirow{2}{*}{\textbf{Model}} &\multicolumn{4}{c}{\textbf{Twitter-gender}}&\multicolumn{4}{c}{\textbf{Twitter-location}}
&\multicolumn{4}{c}{\textbf{Blog-gender}} &\multicolumn{4}{c}{\textbf{Blog-age}}\\ \cmidrule(lr){2-5} \cmidrule(lr){6-9} \cmidrule(lr){10-13} \cmidrule(lr){14-17} &
         BiLSTM & M-GRU & ConvNets & C-LSTM  & BiLSTM & M-GRU & ConvNets & C-LSTM&BiLSTM & M-GRU & ConvNets & C-LSTM  & BiLSTM & M-GRU & ConvNets & C-LSTM\\
         \midrule
          BiLSTM & 93.65 &42.86&50.76 &47.62&
          96.53&36.32&38.95&36.84&87.09&72.00 &70.67 &68.00 & 100.00&56.58 &39.47 & 43.42  \\
          M-GRU &34.62& 88.46 &65.38&46.51&30.77& 92.31 &69.23&38.46&76.67 &83.33&63.33 &56.67 &67.57 &86.49&72.97 &59.46 \\
          ConvNets &51.72&55.17& 89.66 &58.62&39.13&37.50&85.71 &43.49&61.26 &53.33 &90.77&59.26 &48.65 &31.03 &81.08&62.16  \\
          C-LSTM &36.36&33.33&42.42& 90.91 &38.24&35.29&47.06& 88.24 &67.86 &60.71 &57.14 & 89.79&60.53 & 32.05 &39.84 &85.63\\
    \bottomrule
    \end{tabular}
\end{table*}

\subsection{Adversarial Training}\label{subsec:Adv_training}
As aforementioned, attribute inference attackers may detect adversarial examples or defenses in place and train more robust models to evade such protection and thus enhance the inference accuracy. In this respect, we investigate a more robust target model based on adversarial training, which is considered as one of the most empirically effective ways to improve the model robustness against adversarial attacks \cite{goodfellow2014explaining}, to further evaluate the effectiveness of Adv4SG under this setting. More specifically, in this part we study if adversarial training can strengthen the inference attack and lower the success rate of our defense method. We use Adv4SG to generate adversarial texts from random 50\% of correctly classified training data, and incorporate these crafted adversarial examples into the training process, with which, we retrain the BiLSTM inference model under the same parameter setting described in Section~\ref{subsec:experimentalsetup}. Afterwards, we follow the same paradigm to perform Adv4SG over adversarially trained models to test the success rate under four different attribute inference tasks. 

The results are illustrated in Table~\ref{tab:adversarialtraining}. From our results, we can observe that adversarial training barely improves the robustness of inference models against our adversarial attack Adv4SG. The updated success rates of Adv4SG over the inference models after adversarial training are 97.40\%, 71.82\%, 75.95\% and 89.87\% on Twitter-location, Twitter-gender, blog-age, and blog-gender, respectively, which yield no significant difference from the success rates over the original models. These results demonstrate the resilience of the perturbations generated by Adv4SG and the difficulty for inference attackers in defending against our adversarial attack. On the other hand, the relatively weak learning ability of the inference model we deploy in our experiments may somewhat contribute to the success of Adv4SG. This inspires our future work in increasing the learning robustness and capability of NLP models and the advance of adversarial attacks against them.

\begin{table}[t]
    \centering
    \footnotesize
    \caption{Success rates on models with (Adv\_model) and without adversarial training (Ori\_model)}\label{tab:adversarialtraining}
    \tabcolsep=2.8pt
    \begin{tabular}{lcccc}
    \toprule
         \textbf{Model} & \textbf{Twitter-location} & \textbf{Twitter-gender} &\textbf{Blog-age} & \textbf{Blog-gender} \\
         \midrule
          Ori\_model & 97.40\% & 74.03\% &82.28\% &88.61\%\\
          Adv\_model & 97.40\% & 71.82\% &75.95\% &89.87\% \\
    \bottomrule
    \end{tabular}
\end{table}

\section{Applicability and Limitations}\label{sec:limitations}

\begin{figure}[t]
	\centering
	\includegraphics[width=0.5\linewidth]{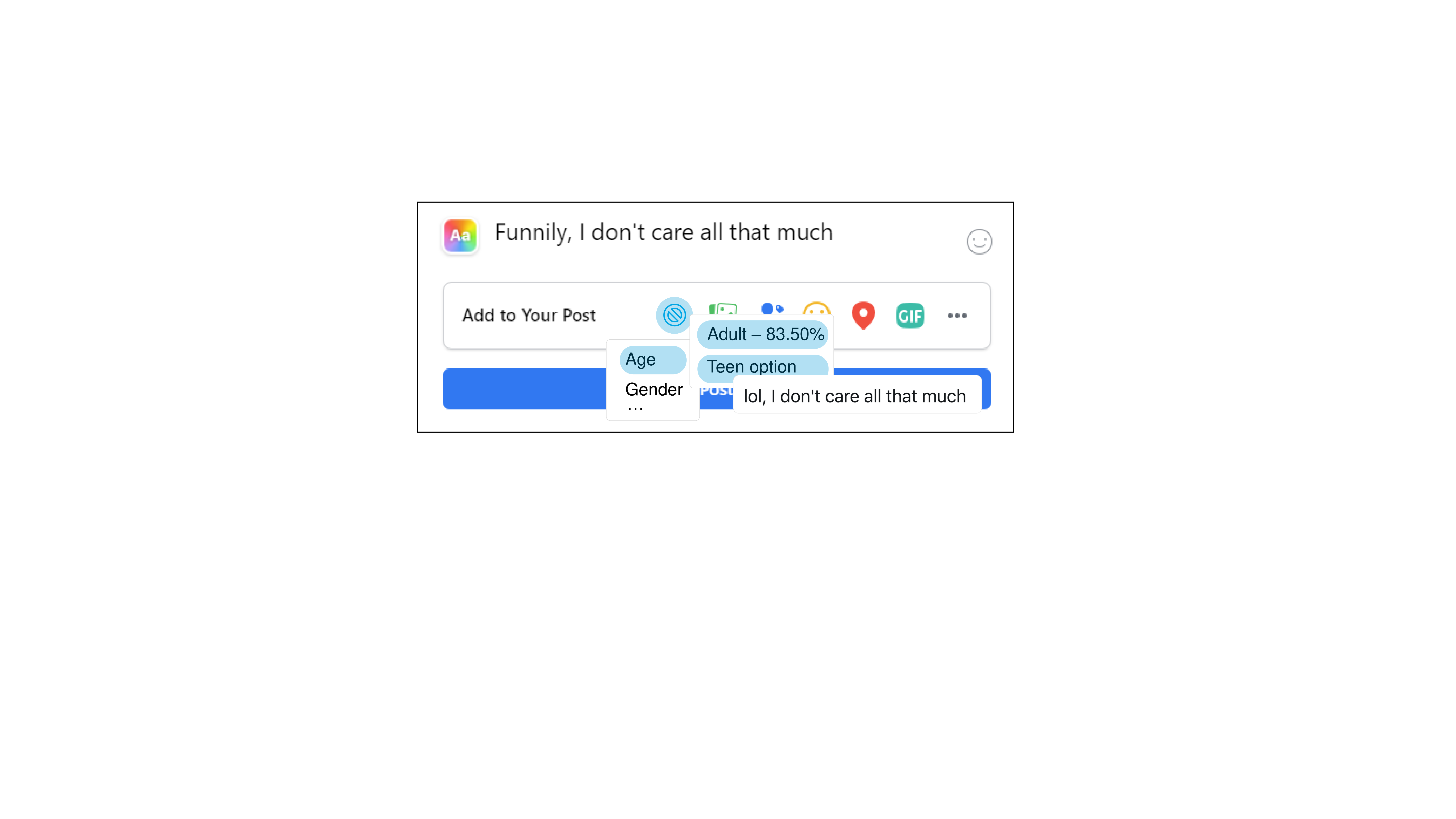}\\
	\caption{A conceptual example of attribute obfuscation service.} \label{fig:service}
\end{figure}

For its applicability, Adv4SG should be an easy-to-use service provided on users' social media client side, so that its privacy protection functionality would be realized in practice. For example, Adv4SG can be developed as an API that is integrated into social media posting and editing systems to allow users to choose the adversarial text according to their provided attribute and text content. A conceptual example of such an attribute obfuscation service devised in Facebook is illustrated in Fig.~\ref{fig:service}, which can change the private attribute that people are unwilling to disclose (i.e., age) of a post to wrong results. Once users give privileges to this adversarial perturbation, the posting data will be obfuscated and updated on behalf of the users. Although not all users might consistently accept the obfuscation feature, we think the possibility of conveniently and proactively perturbing public data can also promisingly increase the uncertainty and difficulty to the attackers. Similarly, our designed method Adv4SG can serve to exhaustively obfuscate the social media data before making it publicly available. 

Nonetheless, our approach also poses some challenges and limitations which we discuss as follows. (1) We successfully perform Adv4SG over the annotated public data in this work, while the real social media lacks the ground truth, which disables Adv4SG from generating the adversarial texts in a real-time fashion. To better obfuscate the attributes, we may need to first recognize the target attribute labels. Though attribute recognition is irrelevant for the scope of our work, it is an interesting supplement to perform few-shot attribute recognition on limited data, and leverage labeled data for better protection solutions. (2) In our experiments, we simply train some regular attack models for attribute inferences. Though Adv4SG has been validated to be transferable and resilient against adversarial training, the attackers could take advantage of more advanced and robust learning models (e.g., text-graph learning) to infer attributes and thus deteriorate Adv4SG. The investigation on this arms race between text-space adversarial attacks and attribute inferences needs to be further extended in the future work, where advanced and robust models could always be evaded by more complicated and sophisticated adversarial techniques. (3) We successfully reduce the computational cost of Adv4SG by significant 36.85\% on average for different inference tasks from Genetic attack, but it still takes some time to generate the adversarial texts, which is less efficient than Greedy attack and PSO attack and its efficiency on population-based optimization needs to be further improved to support the real-world social media with very large and active user engagements. In addition to using word importance to facilitate text mutations, the local best positions and global best positions used by PSO attack for exploration and exploitation provide some good inspirations to potentially help advance the operations of population sampling and crossover during our adversarial text generation. We leave it as our future work.

We acknowledge these challenges and limitations, yet they do not impact the great value and general validity of our new insight to turn the adversarial attacks into attribute obfuscation and privacy protection in the practical social media environment. 

\section{Related Work}\label{sec:related}
Inference attacks on attributes such as gender, political views, and religious views have been studied in decades \cite{kim2012multiplicative,gong2018attribute}. To protect the user-oriented private data, various protection techniques have been proposed to mitigate such inference attacks. As the most traditional method, anonymizations \cite{liu2008towards,zhou2008preserving,dwork2008differential,yuan2010personalized,andreou2017identity,shetty2018a4nt,beigi2018securing} are developed to either remove or mask the identifiable information on social media, while they are still vulnerable to specific types of data leakage \cite{machanavajjhala2007diversity,li2007t}. Some works focus on obfuscating users' interactions by studying the relationship between privacy and utility to hide their actual intentions and prevent profiling \cite{puglisi2015content,rebollo2010optimized}. Unfortunately, developed machine learning based inferences \cite{oh2017adversarial,gong2018attribute} can easily utilize non-anonymous data to re-identify users. Regarding to this issue, some promising defense methods have been thus presented, such as differential privacy and its variant local differential
privacy \cite{wang2017locally,erlingsson2014rappor,bassily2015local}, deep data obfuscation \cite{keswani2016author,karadzhov2017case}, and game-theoretic optimization \cite{shokri2015privacy,jia2018attriguard,shokri2012protecting}. But they are still suffering from limitations of either cost-expensive, large utility loss, or introducing additional privacy concerns. 

Recently, due to the vulnerabilities of machine learning, adversarial attacks are starting to be leveraged as defensive mechanisms against inference attacks \cite{jia2018attriguard,oh2017adversarial,shetty2018a4nt,jia2019memguard,kumar2020adversary}, which have delivered great potentials. However, most of these works focus on the specific application scenarios where their targets are limited to continuous and high-dimensional space. The investigation into more challenging social media environment and the corresponding data of unstructured discrete property has been scarce. The exception is that Shetty et al. \cite{shetty2018a4nt} exploited generative adversarial networks (GAN) to generate text-space adversarial examples to evade authorship identification, which is suffering from trial-and-error on optimization and hence computationally intractable to provide a realistic solution over large data on social media. 

The existing text-space adversarial attacks \cite{zhang2020adversarial} either borrow the gradient-based optimization routine from image domain that computes perturbations over the embedding space \cite{miyato2016adversarial,samanta2017towards,liang2017deep} or leveraging heuristics to search perturbations from an end-to-end basis in large space \cite{ebrahimi2017hotflip,alzantot2018generating,gao2018black,li2018textbugger}. 
For example, Papernot et al. \cite{papernot2016crafting} generated adversarial texts by using Jacobian matrix, and Sun et al. \cite{sun2018identify} followed C\&W method \cite{carlini2017towards} to migrate adversarial attacks on texts. Both of these attacks are performed on word embeddings, which cannot work in an end-to-end manner. The works presented in \cite{liang2017deep,samanta2018generating} adopted the concept of image-based adversarial attacks that use the cost gradient to identify interesting characters or words. AdvGen~\cite{cheng2019robust} also conducts on the gradient base but it considers the similarity between the loss function's gradient, and the distance between words. TextBugger \cite{li2018textbugger} scores the word importance by computing the Jacobian matrix for the given input text to facilitate greedy token selection, but proceeds by substituting the selected words with the optimal bug from candidates, including similar words in embedding space and word transformations. However, all of these methods are designed under the white-box attacks that are lack of practicability in real-world application scenarios where attackers may know nothing about the target models, or cannot access model structure and parameters. Under black-box attack setting, Jia et al. \cite{jia2017adversarial} and Wang et al. \cite{wang2018robust} constructed adversarial examples by adding meaningless sentences to the texts. Gao et al. \cite{gao2018black} designed the attack WordBug that scores word importance by removing it from text, and perturbs words in the descending order regarding word importance scores using character transformations. Alzantot et al. \cite{alzantot2018generating} used genetic optimization algorithm to generate adversarial examples with semantically similar candidates, where population sampling is performed in a random way at each generation. To further improve the attack effectiveness and efficiency, Zang et al. \cite{zang2019word} elaborated sememe-based annotation method to generate word's substitutions and adapted particle swarm optimization (PSO) strategy to expedite best position search for adversarial examples. In addition, there are also some other recent works \cite{sato2018interpretable,zhang2020generating,ren2019generating,samanta2017towards} that contribute to either word substitution candidate preparations, or perturbation optimization for adversarial text generations. All these attacks indicate that word-level perturbations perform comparatively better from the perspectives of attack efficiency and adversarial example quality. 

In this paper, we study the applicability of text-based adversarial attacks on social media and investigate how to adapt adversarial attacks for social-good applications. In a recent work, Li et al. \cite{li2021turning} proposed a text-space adversarial attack for social media privacy protection by formulating new candidate construction and optimization procedure. In this work, we focus on the similar problem but design an upgraded practical method. First, we consider the more challenging black-box scenario where we don’t rely on any knowledge of the threat model, even the query results during the optimization. Besides, We propose a comprehensive method by integrating the gradient information and perturbation variance to more accurately find important word tokens to perturb, which jointly guarantee the success rate and cost efficiency.




\section{Conclusion}\label{sec:conclusion}

In this paper, we investigate adversary for social good, and cast attribute privacy protection problem on social media as an adversarial attack formulation problem to defend against attribute inference attacks. We focus on text data in our problem and propose a text-space adversarial attack Adv4SG under the black-box setting, where the attack constraints are first defined; guided by them, a sequence of plausible perturbations are automatically performed to generate the adversarial texts using semantically and visually similar word candidates, which are regulated by a reformed population-based optimization algorithm. We conduct comprehensive experimental studies on real-world social media datasets to evaluate the performance of Adv4SG, which validate its effectiveness and efficiency against attribute inference attacks. Despite the challenges and limitations, we believe that our work unveils novel insight of turning adversarial attacks in machine learning into defense strategies and implies the great potential on the applicability of adversarial attacks for attribute obfuscation and privacy protection in practice.







\bibliographystyle{ACM-Reference-Format}
\bibliography{main}


\end{document}